\documentclass[acmsmall]{acmart}
\usepackage{graphicx}
\usepackage{booktabs}
\usepackage{multirow}
\usepackage{array}
\usepackage{hyperref}
\usepackage{url}
\usepackage{xcolor}
\usepackage{listings}
\usepackage{mdframed}
\usepackage{algorithm}
\usepackage{algpseudocode}

\newmdenv[
  linewidth=0.8pt,
  roundcorner=4pt,
  skipabove=8pt,
  skipbelow=8pt,
  innerleftmargin=10pt,
  innerrightmargin=10pt,
  innertopmargin=8pt,
  innerbottommargin=8pt
]{examplebox}

\theoremstyle{definition}

\theoremstyle{remark}

\title{Inspire or Predict? Exploring New Paradigms in Assisting Classical Planners with Large Language Models}

\author{Wenkai Yu}
\affiliation{
  \institution{State Key Laboratory of Public Big Data, Guizhou University}
  \country{China}
}
\email{gs.wkyu24@gzu.edu.cn}

\author{Jianhang Tang}
\affiliation{
  \institution{State Key Laboratory of Public Big Data, Guizhou University}
  \country{China}
}
\email{jhtang@gzu.edu.cn}

\author{Yang Zhang}
\affiliation{
  \institution{College of Computer Science and Technology, Nanjing University of Aeronautics and Astronautics}
  \country{China}
}
\email{yangzhang@nuaa.edu.cn}

\author{Yixiong Feng}
\affiliation{
  \institution{School of Mechanical Engineering, Guizhou University}
  \country{China}
}
\email{fyxtv@zju.edu.cn}

\author{Celimuge Wu}
\affiliation{
  \institution{Meta-Networking Research Center, The University of Electro-Communications}
  \country{Japan}
}
\email{celimuge@uec.ac.jp}

\author{Kebing Jin}
\authornote{Corresponding author.}
\affiliation{
  \institution{State Key Laboratory of Public Big Data, Guizhou University}
  \country{China}
}
\email{kbjin@gzu.edu.cn}

\author{Hankz Hankui Zhuo}
\affiliation{
  \institution{School of Artificial Intelligence, Nanjing University}
  \country{China}
}
\email{hankz@nju.edu.cn}

\begin{document}
% =========================
% Abstract
% =========================
\begin{abstract}
Addressing large-scale planning problems has become one of the central challenges in the planning community, deriving from the state-space explosion caused by growing objects and actions. Recently, researchers have explored the effectiveness of leveraging Large Language Models (LLMs) to generate helpful actions and states to prune the search space. However, prior works have largely overlooked integrating LLMs with domain-specific knowledge to ensure valid plans. 
In this paper, we propose a novel LLM-assisted planner integrated with problem decomposition, which first decomposes large planning problems into multiple simpler sub-tasks with dependency construction and conflict detection. Then we explore two novel paradigms to utilize LLMs, i.e., LLM4Inspire and LLM4Predict, to assist problem decomposition, where LLM4Inspire provides heuristic guidance according to general knowledge and LLM4Predict employs domain-specific knowledge to infer intermediate conditions. We empirically validate the effectiveness of our planner across multiple domains, demonstrating the ability of search space partition when solving large-scale planning problems. The experimental results show that LLMs effectively locate feasible solutions when pruning the search space, where infusing domain-specific knowledge into LLMs, i.e., LLM4Predict, holds particular promise compared with LLM4Inspire, which offers general knowledge within LLMs. The source code is available at \url{https://github.com/isinawa/LLMs-assisted_planning/tree/master}.

\end{abstract}

\maketitle

% % =========================
% % Keywords
% % =========================
% \begin{keyword}
% Planing with LLMs \sep Divide and rule \sep Classic planner \sep PDDL
% \end{keyword}
% \end{frontmatter}

%%%%%%%%%以下为正文

\section{Introduction}
Planning aims to generate a course of actions or policies that transform given initial states into goal states, provided that domain models are available, either designed by experts or learned from training data or interactions with the environment. However, as the numbers of objects and actions increase, traditional planning methods often struggle to find high-quality solutions within an acceptable time, which limits the real-world applicability of planning techniques.

Classical planning methods \cite{blum1997fast,lavalle2006planning} have demonstrated strong performance in finding feasible plans for required goals. However, the increasing scale of planning problems leads to exponential growth in the search space, resulting in state-space explosion and consequently limiting planning efficiency and solution quality \cite{ghallab2014automated,helmert2006fast,bonet2001planning}. A natural way to address large-scale planning is either to prune unpromising regions of the search space or to decompose a complex problem into several simpler subproblems that can be solved separately \cite{ghallab2014automated,helmert2006fast,bonet2001planning,hoffmann2001ff}. For example, Fast Downward constructs and exploits rich structural representations, such as domain transition graphs and causal graphs, to expose hidden constraints and guide search more effectively \cite{helmert2006fast}. Goal-ordering and goal-agenda methods reduce complexity by solving suitably ordered subsets of goals incrementally, while sub-goal partition approaches further divide a problem into loosely coupled components \cite{koehler2000reasonable,chen2004sgplan}.
Nevertheless, those approaches can be costly in large-scale settings with huge object scales and complex constraints when analyzing the problems and constructing structural representations to indicate the relations between objects. Recent work on sketches also improves efficiency by explicitly exploiting sub-goal structure to guide planning toward simpler and more manageable subproblems \cite{drexler2024expressing}. However, that approach requires historical traces to learn sketches for decomposition, and therefore lacks general strategies when facing unseen domains without prior experience. 
In summary, when facing large-scale problems, the above methods often incur substantial overhead due to their reliance on constructing structured representations to capture dependency relations. Some of them even require manually defined rules or the learning of historical trajectories to improve planning efficiency. Moreover, these methods are typically complex and self-contained systems, which makes it difficult for them to incorporate guidance from external components. To address this issue, we propose a simple decomposition mechanism that incrementally simplifies the problem by constructing dependency structures, while allowing external components to be integrated during planning to dynamically infer decomposition strategies \cite{kambhampati2024position,valmeekam2023planning,jin2025integrating}.

Meanwhile, inspired by the strong inference capability of Large Language Models (LLMs), recent work has explored using LLMs as policies by directly querying them based on historical observations and actions \cite{li2022pre,huang2022language,ahn2022can,huang2022inner,brohan2022rt,zitkovich2023rt}. According to their built-in common-sense knowledge, LLMs have also been used to assist planners as external components \cite{paulius2024bootstrapping,liu2023llm,wei2025plangenllms}, for example, as heuristic guides, rather than as fully independent planners, due to their lack of domain-specific logical constraints \cite{valmeekam2023planning,kambhampati2024position,valmeekam2024llms}. However, without explicit domain-specific constraints, LLMs cannot guarantee that their predictions are feasible or executable \cite{zhu2025knowagent}. Therefore, in this paper, by comparing LLM-assisted planning based on general knowledge with that constrained by domain-specific knowledge, we aim to explore the following question: \textbf{What roles can LLMs play within planning frameworks, and which one shows greater promise for LLM-assisted planning?}

\begin{figure}[h!]
    \centering
    \includegraphics[width=0.9\textwidth]{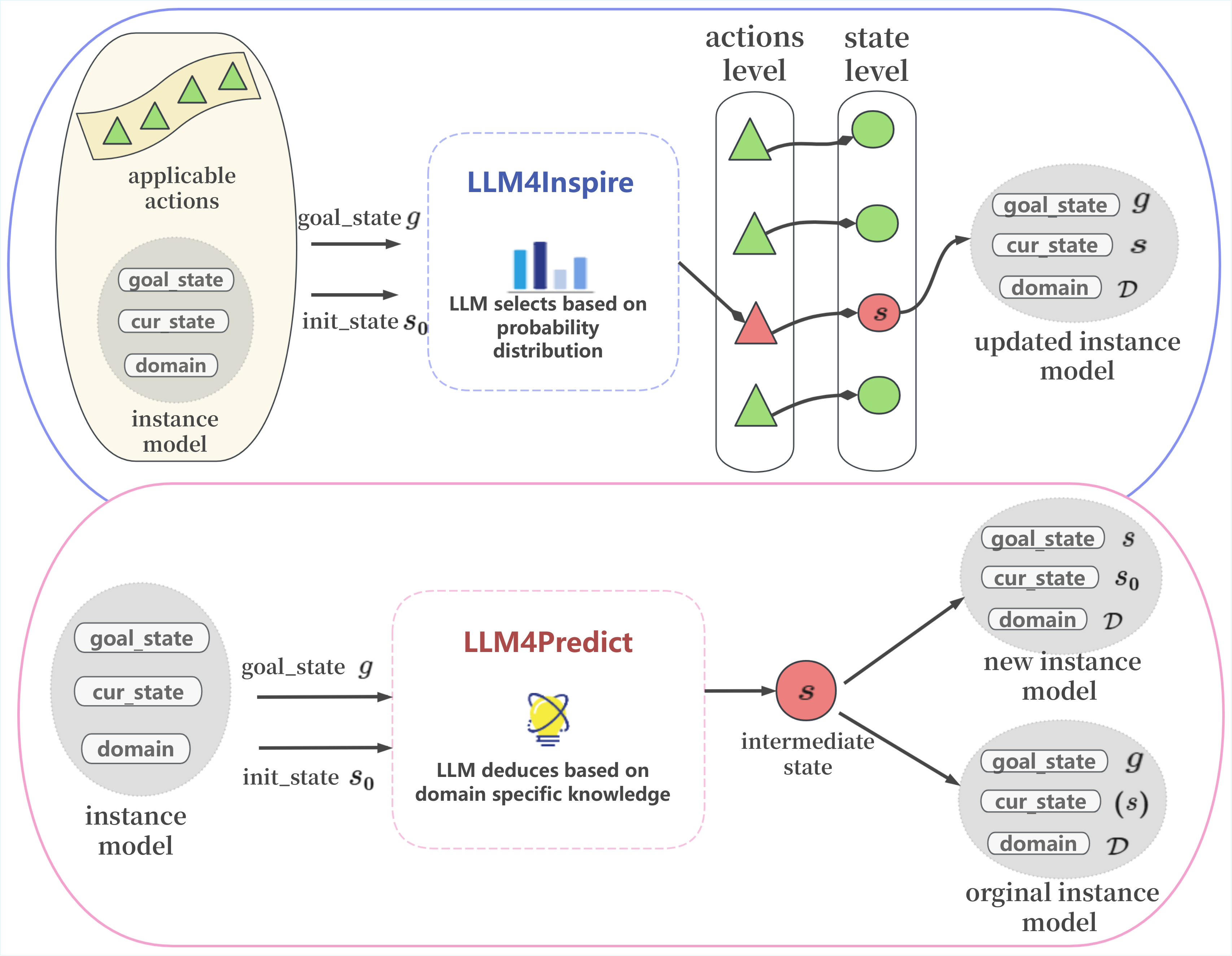}
    \caption{Two paradigms of utilizing LLMs in planning.}
    \label{fig:Overview}
\end{figure}

To address large search spaces while leveraging LLMs for planning assistance, we propose a decomposition-based planner with two distinct paradigms for integrating LLMs into planning frameworks: \textit{LLM4Inspire} and \textit{LLM4Predict}. Specifically, since feasible plans often depend on the order in which goals are achieved, we first employ a directed acyclic graph to decompose the original large-scale problem into subproblems. We then leverage LLMs to further assist decomposition through two different mechanisms. As shown in Figure \ref{fig:Overview}, \textit{LLM4Inspire} uses LLM-assisted heuristics to select valid actions toward goals from the set of applicable actions, whereas \textit{LLM4Predict} predicts intermediate states between the current state and the goal state to further prune the search space. Based on the resulting sub-tasks, we use an existing planner to solve them sequentially and compose the final plan. Through comparing these two LLM-assisted planning paradigms, this paper investigates whether the built-in general knowledge of LLMs can substitute for domain-specific knowledge during planning inference. Experimental results show that predicting intermediate states to partition the search space, as in \textit{LLM4Predict}, performs better than relying on common-sense action guidance, as in \textit{LLM4Inspire}, highlighting the importance and non-substitutability of domain-specific constraints.

The remainder of the paper is organized as follows. We first introduce related work and formally define the planning problem. We then present our approach in detail and evaluate it against representative baselines. Finally, we conclude the paper and discuss future work.

\section{Related Work}
\textbf{Classical planning methods.} Classical planning has achieved remarkable success in domains with explicit structural rules. Heuristic search and graph-based planning methods perform effectively on problems of moderate scale \cite{blum1997fast,helmert2006fast,bonet2001planning,hoffmann2001ff}, but their efficiency often declines sharply as the number of objects, goals, and interactions increases. To improve scalability, divide-and-conquer strategies have been widely explored in planning, especially in hierarchical approaches such as Hierarchical Task Networks (HTN), which decompose a complex task into smaller and more manageable sub-tasks \cite{nau2003shop2,alford2012htn}. These methods have shown strong performance in applications such as logistics and rescue operations \cite{nau2003shop2,alford2012htn}. The idea of decomposition in classical planning also inspires recent efforts to exploit subgoal structure more explicitly \cite{drexler2024expressing}, and provides important motivation for addressing large-scale planning problems.

\textbf{LLMs for planning.} With the rapid development of large language models, many recent studies have investigated their use in planning. Representative approaches include chain-of-thought \cite{wei2022chain}, tree-of-thought\cite{yao2023tree}, and zero-shot planners\cite{huang2022language}, which guide LLMs to generate action sequences through prompting. Other studies further combine LLMs with external validators or planning tools, so that candidate plans generated by LLMs can be checked and refined through symbolic feedback \cite{liu2023llm,valmeekam2023planning,kambhampati2024position}. These works demonstrate that LLMs possess useful world knowledge and can provide helpful guidance for planning \cite{li2022pre,hao2023reasoning}. However, recent studies also point out that their inference ability remains limited, especially in unfamiliar domains, and that directly relying on LLMs to produce executable and complete plans is still unreliable \cite{wei2025plangenllms,valmeekam2023planning,kambhampati2024position,valmeekam2024llms}. Therefore, although LLMs are promising for planning, they are better viewed as auxiliary components rather than fully independent planners\cite{kambhampati2024position}.

\textbf{LLMs as components in planning frameworks.} Recent research has increasingly shifted from treating LLMs as end-to-end planners to integrating them into existing planning frameworks. Survey and perspective studies in automated planning and scheduling suggest that the most promising direction is not to replace symbolic planning entirely, but to combine the generative capability of LLMs with the reliability of existing planning machinery \cite{kambhampati2024position,pallagani2024prospects}. In this line of work, LLMs have been used as object-level planning aids \cite{paulius2024bootstrapping}, knowledge-augmented planning components \cite{zhu2025knowagent}, world models for reasoning and planning \cite{hao2023reasoning}, or guides embedded into structured search procedures such as Monte Carlo Tree Search \cite{zhao2023large}. These methods show that LLMs can improve planning efficiency when coupled with external reasoning mechanisms \cite{paulius2024bootstrapping,zhu2025knowagent,zhao2023large}. Our work is related to this trend, but differs from it in an important way. Rather than using LLMs as direct planners, our approach embeds them into a conflict-aware decomposition-and-composition framework built on a verifiable planner. Unlike existing studies that mainly use LLMs for direct plan generation, formalization, or search guidance, our method focuses on large-scale planning through conflict-aware decomposition and composition within a verifiable symbolic planning framework.

\section{Problem Formulation}

In this paper, we aim to solve classical planning problems, where a classical planning problem can be formally defined as a triple $P = \langle s_0, g, D\rangle$. $s_0$ denotes the initial state, represented as a set of ground atoms (propositions) that characterize the facts holding in the initial configuration. $g$ denotes the goal specification, defined as a set of ground atoms that must be satisfied in any solution state. $\mathcal{D}$ denotes the domain model, comprising a finite set of action models. Each action model is a quadruple $\mathcal{A} = <a, \mathit{pre}(a), \mathit{add}(a), \mathit{del}(a)>$, where $a$ is an action name with zero or multiple parameters. $\mathit{pre}(a)$ is a precondition set indicating the conditions under which it can be applied. $\mathit{add}(a)$ and $\mathit{del}(a)$ are respectively an adding and deleting list to form the effects of the action
$a$. Each action $a$ is characterized by its preconditions $\mathit{pre}(a)$, add effects $\mathit{add}(a)$, and delete effects $\mathit{del}(a)$. An action $a$ is applicable in a state $s$ if and only if $\mathit{pre}(a) \subseteq s$. When applied, $a$ transforms $s$ into a new state by adding all atoms in $\mathit{add}(a)$ and removing all atoms in $\mathit{del}(a)$.

The aim is to generate a plan $p$, i.e., a grounded action sequence, to achieve goals $g$ from the initial state $s_0$.

\begin{examplebox}

To have a clearer understanding, consider a simplified Blocks domain with three blocks $A$, $B$, and $C$.

Initial state $s_0$: \texttt{ontable(A)}, \texttt{ontable(B)}, \texttt{ontable(C)}, \texttt{clear(A)}, \texttt{clear(B)}, and \texttt{clear(C)} indicate that $A$, $B$, and $C$ are on the table and not stacked.

Goal state $g$: \texttt{on(A,B)}, \texttt{on(B,C)} indicate that the goal is to put $A$ on $B$, and put $B$ on $C$ 

Plan $p$: $<$\texttt{pick-up(B)}, \texttt{stack(B,C)}, \texttt{pick-up(A)}, \texttt{stack(A,B)}$>$ indicates that we pick up $B$ from the table, place it on $C$,  pick up $A$, and place it on $B$. When executed starting from $s_0$, leads to a state satisfying $g$:
\begin{enumerate}
    \item Move block $B$ from the table and place it on block $C$.
    \item Move block $A$ from the table and place it on block $B$.
\end{enumerate}

This sequence of actions transforms the initial state $s_0$ into the goal state $g$, clearly illustrating the components and solution process of a classical planning problem.
\end{examplebox}

\section{LLM-assisted Planning with Problem Decomposition}
In this section, we address our LLM-assisted planner for complex planning problems in detail. When tackling complex problems, classical planners face several major challenges: (1) When the problem involves a large number of objects and the state space becomes exceedingly vast, the number of states that the planner must explore grows exponentially. (2) In addition, when achieving certain predicates that constitute parts of the goal state requires an excessively long sequence of steps, the planner cannot traverse the exceedingly deep search tree within the time constraints. (3) For static heuristic functions, they are unable to precisely leverage comprehensive world models to respond to the planner with dynamic and flexible guidance quickly.

\begin{figure*}[!h]
    \centering
    \includegraphics[width= 0.98\textwidth]{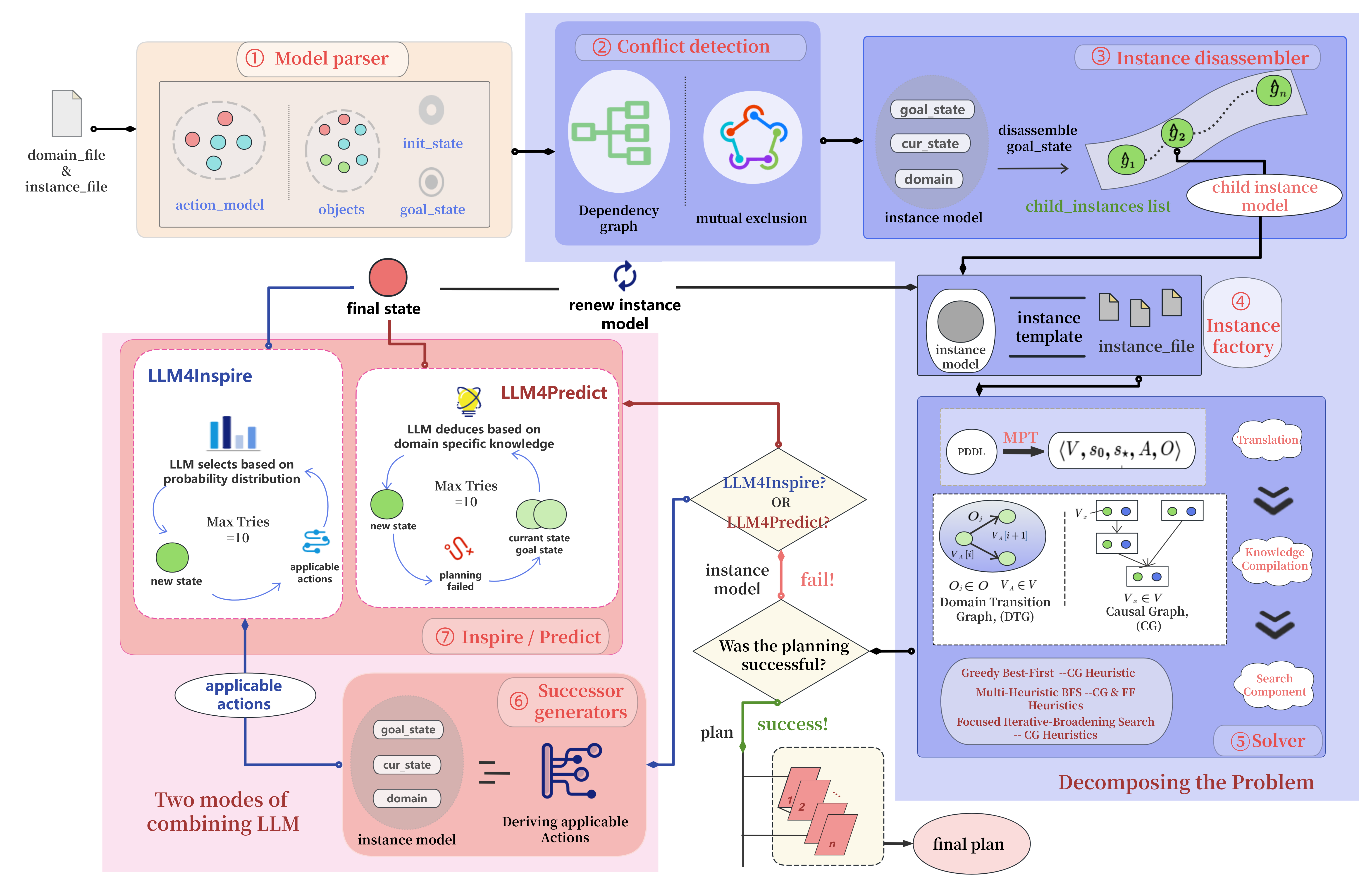}
    \caption{Two modes of combining LLMs with traditional planners}
    \label{fig:Inspire/Predict}
\end{figure*}

To address those issues, we propose a novel LLM-assisted planner integrated with problem decomposition, as shown in Figure \ref{fig:Inspire/Predict}, comprising seven core modules: Model Parser, Conflict Detection, Instance Disassembler, Instance Factory, Solver, Successor Generators, and LLM-assisted Modules (Inspire / Predict):\\
1. \textbf{Model Parser} takes domain and instance files as inputs and outputs instance models for planners.\\
2. \textbf{Conflict Detection} constructs dependency graphs to detect conflicts, eliminates them by computing immediate states, and updates the instance models.\\
% after all mutex conditions are implemented.\\
3. \textbf{Instance Disassembler} receives the instance models, decomposes them into sub-instances based on atomic goal states, and outputs lists of sub-instances.\\
4. \textbf{Instance Factory} converts the sub-instance models into standardized PDDL sub-instance files.\\
5. \textbf{Solver} employs a general planner, the Fast Downward planner (we call it DW) in this paper. \\
6. \textbf{Successor Generator} computes executable action sequences for the current state toward goals. \\
7. \textbf{Inspire / Predict Module} generates landmark states to help planners search actions toward the goals and obtain simpler sub-instance models for planners.

 Specifically, we utilize Model Parser to parse PDDL planning problems and apply Conflict Detection to generate mutex conditions that update the instance model. Next, we decompose the goals of the complex problem into a list of sub-instance models through the Instance Disassembler and Instance Factory, thereby constructing an ordered set of sub-tasks for an existing planner, i.e., Solver, to process. 
 % Next, for decomposed subproblems that the Solver cannot handle, 
 For decomposed subproblems that the solver fails to handle due to timeout, we employ LLMs to generate landmark states, which facilitate the creation of simpler sub-instance models and assist the planner in computing actions to achieve the goals, i.e., Successor Generators and Inspire/Predict. Note that we introduce two ways for LLM-assisted classical planners. At last, we iteratively plan over sub-instances and decompose complex planning problems until the goal state is achieved.

\subsection{Conflict Detection and Problem Decomposition} 

\begin{figure}[h]
    \centering
    \includegraphics[width=0.9\textwidth]{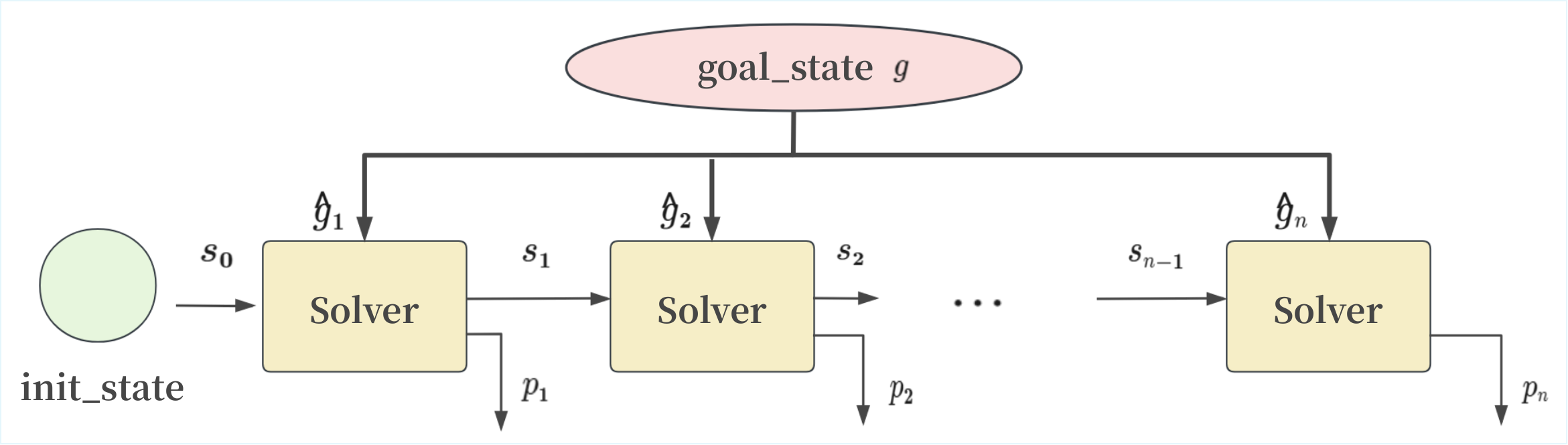}
    \caption{Problem decomposition framework.}
    \label{fig:decomposing_framework}
\end{figure}

To address large-scale planning problems by reducing problem complexity, we propose a divide-and-conquer method that partitions the original large-scale problem into a set of subproblems, as illustrated in Figure \ref{fig:decomposing_framework}. 

However, during dividing the problems, the key difficulty is that generating subproblems according to goals $g$ is insufficient, since ordering constraints may exist and unpredictable conditions need to be considered based on the initial state. 

\textit{For example, in the Blocks domain, if the goal is $g = \{\text{on}(B,C), \text{on}(A,B)\}$, achieving $\text{on}(B,C)$ must precede $\text{on}(A,B)$, since $A$ cannot be placed on $B$ until $B$ is correctly positioned on $C$. Suppose that initial state is $s_0 = \{\text{on}(C,A), \text{ontable}(B), \text{ontable}(A)\}$, if we first achieve $\text{on}(B,C)$ and the updated state will be $s_i = \{\text{on}(C,A), \text{on}(B,C), \text{ontable}(A)\}$. To achieve $\text{on}(A,B)$, we must move $C$ away from $A$, which need undo $\text{on}(B,C)$.} 
%%% 这里实际上缺失了对conflict的定义，解释为什么成环就是conflict。
%1.这个挑战/困难的原因是。。（要提到环就是冲突） 2.由此我们怎么做。。 3.具体实现的过程
The object dependency relationships within each state can be modeled as directed acyclic graphs. The process of sequentially achieving sub-goals from the initial state is equivalent to constructing goal dependency graphs by identifying the nodes in the initial dependency graphs in sequence. During this process, cycles may emerge in the graphs, a phenomenon we refer to as conflicts, regarded as an attempt to access an already occupied node. Therefore, in this paper, we compute immediate states to dismantle the previously established dependency relationships to free up such a node. Specifically, we simulate this process of achieving the ordered sub-goals by constructing dependency graphs to detect the conflicts and generating mutually exclusive conditions to eliminate the conflicts, thereby ensuring that all sub-goals can be achieved in order.
% We then order the goals $g$ based on the graphs and achieve the conditions to assist the ordered goals can be achieved without undo.

%%% 每部分对应哪一个模块需要提及
% %%%%%%%%%%%%%%%%%%%%%%%%%%%%%%%%%%%%%%%%%%%%%%%%%%%
% %%%%%%%%%%%%%%%%%%%%%%%%%%%%%%%%%%%%%%%%%%%%%%%%%%%
% \begin{figure}[!t]
\begin{figure}[h]
    \centering
    \includegraphics[width=0.9\textwidth]{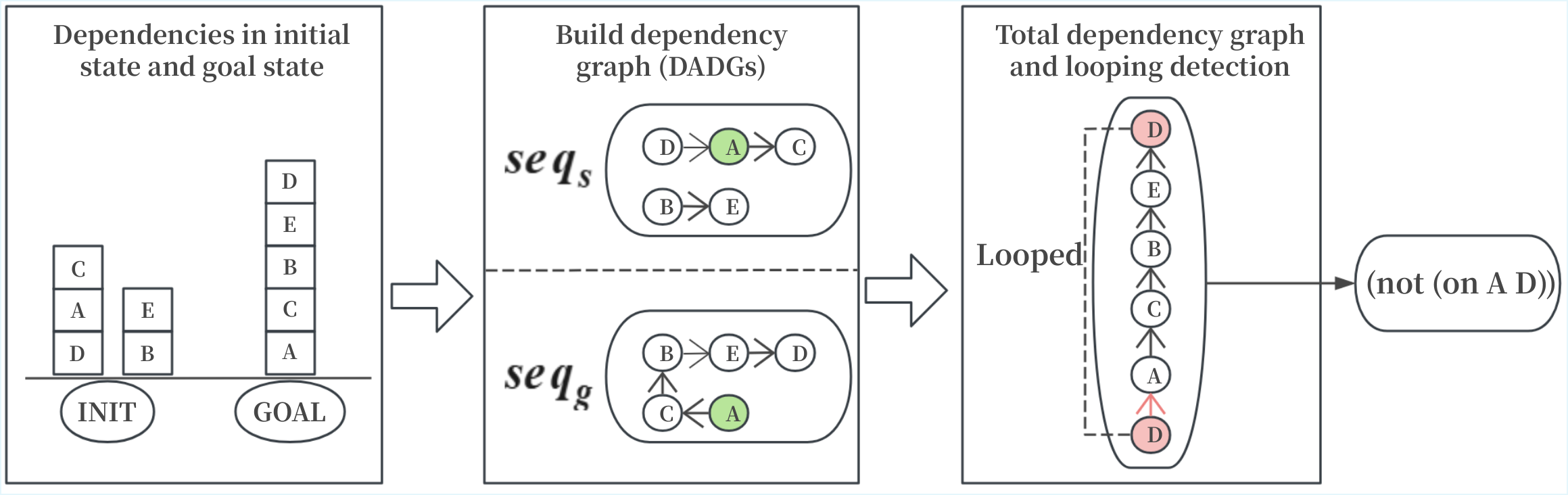}
    \caption{Conflict detection graph.}
    \label{fig:conflict_detection}
\end{figure} 

The following text will explain the implementation details of the Conflict Detection and the Instance Disassembler, respectively. 
For the Conflict detection module, given a planning problem $P=<s_0, g, \mathcal{D}>$, where goals $g=<g_1, g_2, ..., g_n>$.  
We first build directed acyclic dependency graphs (DADGs) $G_g=\{ G^1_g, G^2_g, \dots \}$ and $G_s=\{ G^1_s, G^2_s, \dots \}$ to indicate the dependencies between goals and initial states, respectively. Each node is an object involved in $g$ or $s_0$, and each edge is a predicate, indicating that a dependency relationship exists between the objects. \emph{For example, $\text{on}(A,B)$ is a predicate that indicates object A is on object B. As implementing $\text{on}(A,B)$ changes the state of object B, it implicitly indicates that object B must be stacked before object A, then we create an edge pointing from B to A.}   

%上面讨论完了如何构建依赖图，我们接下来讲述以来序列的构建过程
Based on graphs $G_g$ and $G_s$, we then compute sequences $Seq_g=\{Seq^1_g, Seq^2_g, \dots\}$ and $Seq_s=\{Seq^1_s, Seq^2_s, \dots\}$ for each sub-graph via Topological sorting. Specifically, we first initialize an empty sequence $Seq^i_g = []$ and add the node $Obj^1_i$ into $Seq^i_g$, where $Obj^1_i$ is a node in $G^i_g$ with zero in-degree. Then we add the destination nodes $Obj^k_i$, pointed to by the output edges of $Obj^1_i$, to the $Seq^i_g$. We continuously insert the nodes until all the nodes in the graph $G^i_g$ have been put into $Seq^i_g$. We repeat those procedures until all graphs in $G_g$ have been processed. The way to compute $Seq_s$ is the same, but with $G_s$ as the input.

%上面讲完了依赖序列，接下来我们解释总依赖序列的构建过程以及如何得到互斥条件
To detect the potential conflicts, we explore the cycle formation, which is indicated by the total dependency sequences $T=<T_1, T_2, \dots>$. 
Specifically, we first initialize an empty sequence $T$. 
Next, we take the $Obj^1_i$ in $Seq^i_g$ as the anchor and extract its prefix subsequence $\overline{Seq}_s^k$ from the $Seq^k_s$, where $Obj^1_i \in Seq^k_s$ and $\overline{Seq}_s^k \subset  Seq^k_s$ includes the nodes before $Obj^1_i$. Then, we merge $\overline{Seq}_s^k$ with $Seq^i_g$, i.e., $T_i=[Seq^i_g |\overline{Seq}_s^k]$, and add $T_i$ to $T$. We repeat these steps until all subsequences in $Seq_g$ have been processed. 
%以下是对耦合检测&消除耦合细节的描述
%以下是对耦合检测&消除耦合细节的描述
%以下是对耦合检测&消除耦合细节的描述
After obtaining $T$ completely, we find the conflict by checking whether there is a cycle in $T_i$, and generate the mutually exclusive condition.
Specifically, if any object appears more than twice in $T_i$, we consider that there is a conflict. Based on this conflict, we generate the mutually exclusive condition $\mathcal{C}$ by locating the repeated objects, finding the conflicting edges, i.e., predicates, and negating the predicates. \emph{As shown in Figure \ref{fig:conflict_detection}, object $D$ appears twice in the sequence and points to $A$, so that we regard the edge between $D$ and $A$ as a conflict. As illustrated in the initial state, the edge is $\text{on}(A,D)$ and $\mathcal{C}$ is $\neg \text{on}(A,D)$, accordingly.} 

\begin{algorithm}[h]
\caption{\textbf{Conflict Detection and Problem Decomposition}}
\label{alg:v2}
\begin{algorithmic}[1]
\State \textbf{Input:} Goal state ${g}=<g_1,g_2,..g_n>$, Init state $s_0$
\State \textbf{Output:} Ordered sub-goals $\hat{g}=[\hat{g}_1, \hat{g}_2, \dots, \hat{g}_n]$, Updated init state $s_0$
\State Construct DADGs $G_g$ by $g$ and $G_s$ by $s_0$;
% $\hat{g}=Toposort(G_g), \hat{s} = Toposort(G_s)$;
% \STATE $\hat{T}=Merge(\hat{g}, \hat{s})=<T_1,T_2,..>$
\State Compute $Seq_g$, $Seq_s$ according to $G_g$ and $G_s$;
\State Build $T$ according to $Seq_g$ and $Seq_s$;
\For{$T_i \in T$}
\While{Cycle formation within $T_i$}
\State Generate exclusive condition $\mathcal{C}$, and construct instance $P=<s_0, \mathcal{C}, D>$ based on the Instance Factory Module;
\State Compute plan $p$ for $P$ via the Solver Module;
\State Update state $s_0$ according to plan $p$;
\State Re-calculate $T_i$ base on the state $s_0$;
\EndWhile
\EndFor
\State Compute $\hat{g}$ according to $Seq_g$ and $G_g$;
\State Output $\hat{g}$, $s_0$
\end{algorithmic}
\end{algorithm}

%上面讲完了如何生成互斥条件，接下来我们说明如何根据互斥条件来消除冲突
To eliminate the conflict, we generate action sequences from $s_0$ to $\mathcal{C}$ to create a new state whose goals skip the conflicted nodes. We use the Instance Factory to generate the instance $P = <s_0, \mathcal{C}, D>$, and calculate the plan $p$ via the Solver Module. Based on the $p$, we update the initial state $s_0$ and re-calculate sequence $T_i$ according to $s_0$. We repeat the above steps until the sequence $T$ no longer includes a cycle. 

% %%%%%%%%%%%%%%%%%%%%%%%%%%%%%%%%%%%%%%%%%%%%%%
% We repeat these steps until all subsequences in $Seq_g$ have been processed. Next, we eliminate the conflict by generating the corresponding coupled mutex conditions via detecting the cyclization condition of the dependency sequence $T_i$ and generating the corresponding mutex condition $\mathcal{C}$. %xxx
% %%% 怎么做的
% As shown in Figure \ref{fig:conflict_detection}, the object $D$ in the sequence forms a cycle, and its conflict conditions $(on\ A\ D)$ in the initial state are mutually exclusive from $(not\ (on\ A\ D))$. We use the Instance Factory to generate the instance $P = <s_0, Cond, D>$, and calculate the plan $p$ via the Solver Module. Based on the plan $p$, we update the initial state $s_0$ and re-calculate sequence $T_i$. We repeat the above steps until the sequence $T_i$ no longer forms a cycle. After traversing all the subsequences in $T$, we completely eliminate all conflicts. 
% %%%%%%%%%%%%%%%%%%%%%%%%%%%%%%%%%%%%%%%%%%%%%%

%以上是Conflict Detection 的完整实现过程，接下来是Instance Disassembler
After eliminating all conflicts, we order the sub-goals for future sub-problem generations. We first initialize an empty list $\hat{g}$, traverse each node in $Seq^i_g$ sequentially, and add the corresponding output edges in $G^i_g$ of that node to $\hat{g}$. We repeat the above steps until adding all edges in $G_g$ to $\hat{g}$. Note that the order of $G^i_g$ will not affect the final results, since each object only belongs to one sub-graph $G^i_g$ and no constraints exist between graphs. Finally, we output the sorted goal sequence $\hat{g}$ and the updated initial state $s_0$. 
% The whole procedure is as shown in Appendix A.1, 
The whole procedure is as shown in Algorithm $\ref{alg:v2}$, 
whose target is to ensure that executing all actions from the updated initial state does not undo the resolved sub-goals. After that, we update the instance model, and the Instance Disassembler decomposes the goals into the ordered sub-goals accordingly.

%% 最后对两个模块的执行过程进行总结。
%% 最后对两个模块的执行过程进行总结。
%% 最后对两个模块的执行过程进行总结。
% The purpose of the two modules, Conflict Detection and Instance Disassembler, is to ensure that executing all actions from the updated initial state does not undo the resolved sub-goals. The Conflict Detection detects conflicts via dependency graphs and generates mutex conditions to eliminate conflicts. 
% After all conflicts are eliminated, we update the instance model, and the Instance Disassembler decomposes the goals into the ordered sub-goals based on the updated instance model.

% %%%%%%%%%%%%%%%%%%%%%%%%%%%%%%%%%%%%%%%%%%%%%%%%%%%
% %%%%%%%%%%%%%%%%%%%%%%%%%%%%%%%%%%%%%%%%%%%%%%%%%%%

\subsection{LLM-Assisted Planning Processes}
Although the divide-and-conquer approach indeed reduces the complexity of large-scale planning problems, existing planners may not afford the decomposed search space. In this case, we utilize the comprehensive world knowledge built into LLMs to provide guidance and further divide the unsolvable subproblems. We explore two ways for integrating LLMs into the framework, where LLM4Inspire follows previous research to act as heuristics and LLM4Predict is additionally constrained with domain-specific knowledge to predict future direction.

\subsubsection{Overview of Two LLM-assisted Modes}

\begin{algorithm}[h!]
\caption{\textbf{LLM-assisted planning}}
\label{alg:v1}
\begin{algorithmic}[1]
\State \textbf{Input:} Initial state $s_{0}$, Goal state $g$, Domain $\mathcal{D}$
\State \textbf{Output:} Plan $p$ 
% 在这一块需要加上解耦的描述。
% \STATE Compute DADGs $G$ based on $g$ accordingly ordered sub-goals $\hat{g}=[\hat{g}_1, \hat{g}_2, \dots, \hat{g}_n]$;
% \STATE Build DADGs $G_g$ and $G_s$ from $g$ and $s_0$, decouple to get updated initial state $s_0$ and ordered sub-goals $\hat{g}=[\hat{g}_1, \hat{g}_2, \dots, \hat{g}_n]$.
\State Compute ordered goals $\hat{g}=[\hat{g}_1, \hat{g}_2, \dots, \hat{g}_n]$ and updated state $s_0$ through Conflict Detection and Problem Decomposition.
\State $s$=$s_{0}, p = [~]$; 
\For{$i$ in $n$}
    \State Construct sub-instances $P_{i} = <s,\hat{g}_i,\mathcal{D}>$ based on the Instance Factory Module;
    \State Compute sub-plan $p_{i}$ for $P_i$ by the Solver;
    \If{$p_{i} \neq [~]$}
        \State  $p = [p | p_{i}]$;
    \Else
        \State $times = 0$;
        \While{$p_{i} = [~]$ or $times > 10$}
        \State Utilize LLM4Inspire or LLM4Predict to compute plan $\hat{p}$
        % \IF{Utilize LLM4Inspire}
        %     \STATE Initialize an empty action trajectory $\mathcal{T}$ to record the actions returned by the LLM; 
        %     \STATE Enumerate available actions $\hat{A}$ based on $s$ and $\mathcal{D}$ based on the Successor Generator;
        %     \STATE Generate plan $\hat{p}$ based on $s$, $\hat{g}_i$, $\mathcal{T}$ and $\hat{A}$ via LLM4Inspire, and $\mathcal{T} = [\mathcal{T} |  \hat{p}]$; 
        %     % And $\hat{p}$ will be recorded as a part of $\mathcal{T}$.
        %  \ELSE
        %     \STATE Predict an intermediate state $\tilde{s}$ according to $s$ and $\hat{g}_i$ based on LLM4Predict;
        %     \STATE Construct instance $\hat{P} = <s, \tilde{s}, \mathcal{D}>$; 
        %     \STATE Compute plan $\hat{p}$ for $\hat{P}$ via the Solver;
        % \ENDIF
        \State $p = [p | \hat{p}]$ and update state $s$ according to $\hat{p}$; 
        % \STATE $s$=$\hat{s}$; 
        \State Reconstruct instance $P_{i} = <s, \hat{g}_i, \mathcal{D}>$ and compute sub-plan $p_{i}$ for $P_i$ via the Solver Module, $times$++;
        \EndWhile
        \If{$time>10$ and $p_{i} = [~]$}
            \State return failure;
        \Else
            \State Update state $s$ based on plan ${p_i}$, $p = [p | p_i]$; 
        \EndIf        
    \EndIf
\EndFor
% \STATE $p$, $t$=Integrate(list($P_{1}$,$P_{2}$, ...$P_{n}$))
\State \textbf{Return:} Final plan $p$;
\end{algorithmic}
\end{algorithm}
% \noindent 

As presented in Algorithm~\ref{alg:v1}, we first parse the initial state $s_0$, goal state $g$, and domain models $\mathcal{D}$ by the Model Parser \cite{valmeekam2023planning} (Line 1). Next, we utilize the Conflict Detection to find potential conflict and update the instance model, compute an ordered sub-goals sequence $\hat{g} = [\hat{g}_1, \dots, \hat{g}_n]$ by the Instance Disassembler (Line 3).
% Next, we utilize the Instance Disassembler to construct DADGs $G$ for $g$ and compute an ordered sub-goals sequence $\hat{g} = [\hat{g}_1, \dots, \hat{g}_n]$ through topological sorting (Line 3). 
Then we assign the current state $s$ by $s_0$ and define an empty plan sequence $p$ (Line 4). As for each sub-goal $\hat{g}_i$, we use the Instance Factory to construct a standard sub-instance $P_i$ (Line 6). Then we call an existing planner (Fast Downward in this paper) to solve the problem for a sub-plan $p_i$ (Line 7). If a valid plan exists, we record $p_i$ via $p$ (Lines 8-9). 
Otherwise, we use LLMs to compute plan $\hat{p}$, i.e., LLM4Inspire and LLM4Predict (Line 13). 
% Firstly, for LLM4Inspire, we create an empty action trajectory $\mathcal{T}$ to record the historical actions (Line 14). And we utilize the Successor Generator to compute applicable actions $\hat{A}$ according to $s$ and $\mathcal{D}$ (Line 15). After that, LLMs select the most appropriate action $\hat{p} = [\hat{a}]$ (Line 16). Secondly, LLM4Predict returns an intermediate state $\tilde{s}$ as a temporary goal based on $s$ and $\hat{g}_i$ and utilizes Fast Downward to compute plans $\hat{p}$ according to $<s,\tilde{s},\mathcal{D}>$ (Lines 18-20). 
Based on $\hat{p}$, we update the state $s$ as well as sub-instance $P_i = <s, g_i, \mathcal{D}>$ and utilize Fast Downward to solve $P_i$ (Lines 14-15). We repeated the above procedures at most $10$ times for achieving $g_i$. Nevertheless, we regard $P$ as an unsolvable problem (Lines 17-18). If plan $p_i$ is not empty, we update the state $s_0$ according to $p_i$ and put $p_i$ into $p$ (Line 20). At last, plan $p$ is the final output (Line 24).

\subsubsection{LLM4Inspire}

The subsection introduces the first proposed paradigm of LLMs, i.e., LLM4Inspire, utilizing LLMs to provide applicable actions according to current states. Specifically, the Successor Generator first enumerates all executable actions $\hat{A}$ based on the current state $s$ and domain $\mathcal{D}$. We define a prompt template (please refer to the supplementary) to inform LLMs of available actions, the historical action sequence, the current state, and the goals. Through analyzing possible paths from the current state to the goal state based on their comprehensive knowledge, the LLMs select an optimal action $\hat{p}$ in their eyes. Then we update $s$ by executing $\hat{p}$. 

To obtain the set of executable action sequences, the Successor Generator follows a systematic process. First, it enumerates all grounding actions in the domain model $\mathcal{D}$. Specifically, we first replace all parameter placeholders with objects according to the pre-defined types in the domain models $\mathcal{D}$ for all possible action-object combinations. Then we filter feasible actions $\hat{A} = \{\hat{a}_0, \hat{a}_1, \dots\}$ based on the preconditions and the state $s$, where $pre(\hat{a}_i) \subseteq s$.

\subsubsection{LLM4Predict }
The second paradigm LLM4Predict predicts an intermediate state $\tilde{s}$ between the current state $s_0$ and the goal state $g$. The predicted intermediate state $\tilde{s}$ is constrained by another prompt template (please refer to the supplementary), required to only include a few key predicates. Different from the LLM4Inspire template, LLM4Predict is informed by the current state and goals as constraints, required to generate a state between them. At last, we utilize the Solver Module to solve the intermediate problem $\hat{P} = <s,\tilde{s}, \mathcal{D}>$ and update the state according to the output plan $\hat{p}$ to continuously plan. 

\begin{figure}[h]
    \centering
    \includegraphics[width=0.8\linewidth]{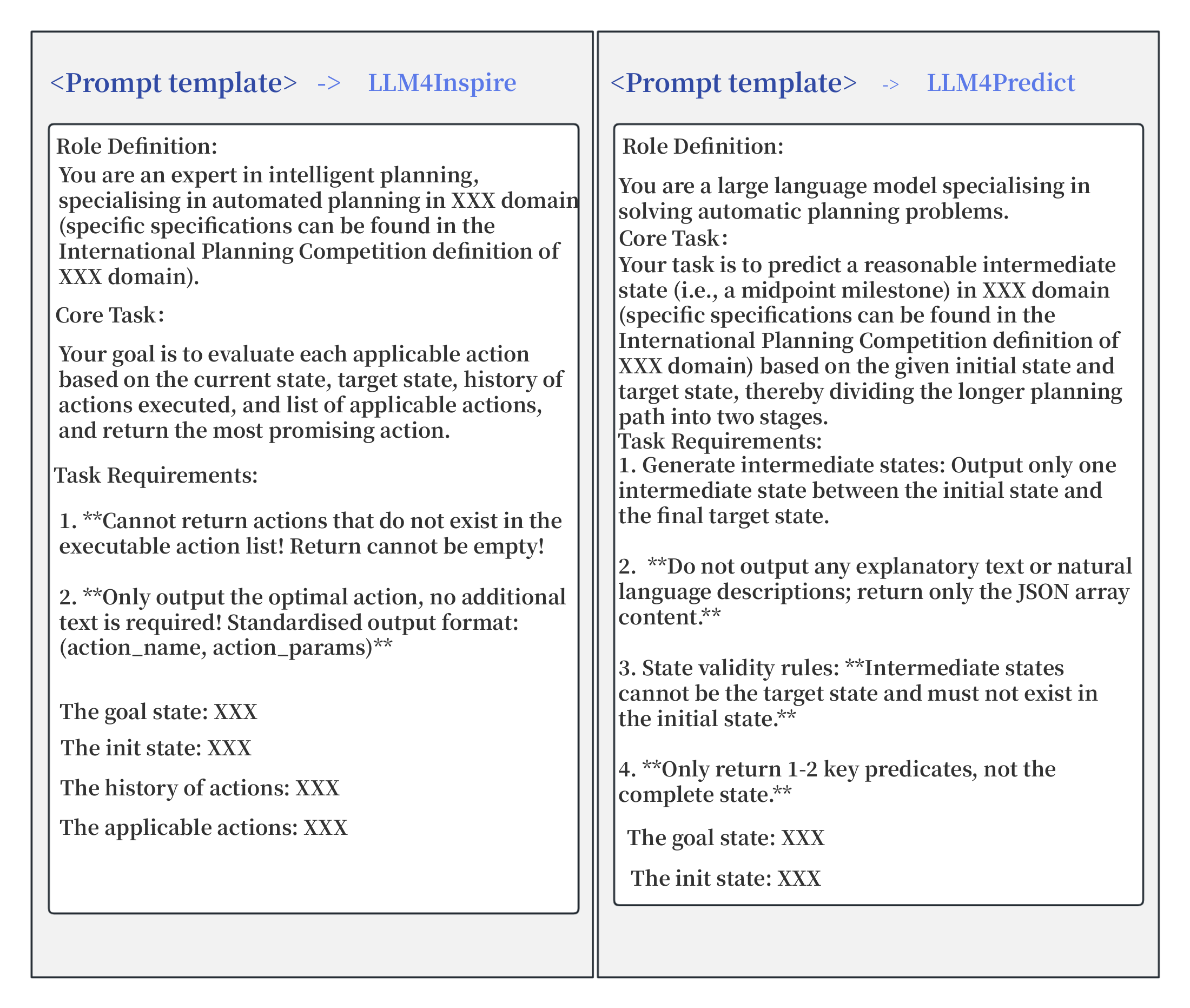}
    \caption{The comparison between prompt templates.}
    \label{fig:prompt}
\end{figure}

\subsection{Analysis and Discussion}

Figure \ref{fig:prompt} illustrates prompt templates of two LLM-based approaches. 
% % % 提示词模板放到附件当中
% The prompt templates of two LLM-based methods are provided in the Appendix A.2.
Both LLM4Inspire and LLM4Predict inform LLMs of the domain name, initial state, and goal state. Differently, LLM4Inspire requires LLMs to evaluate actions from a set of available actions, where LLM4Predict generates an intermediate state instead. 

\subsubsection{The Effectiveness of The Two LLM-based Methods}
\begin{figure}[h]
    \centering
    \includegraphics[width=0.9\textwidth]{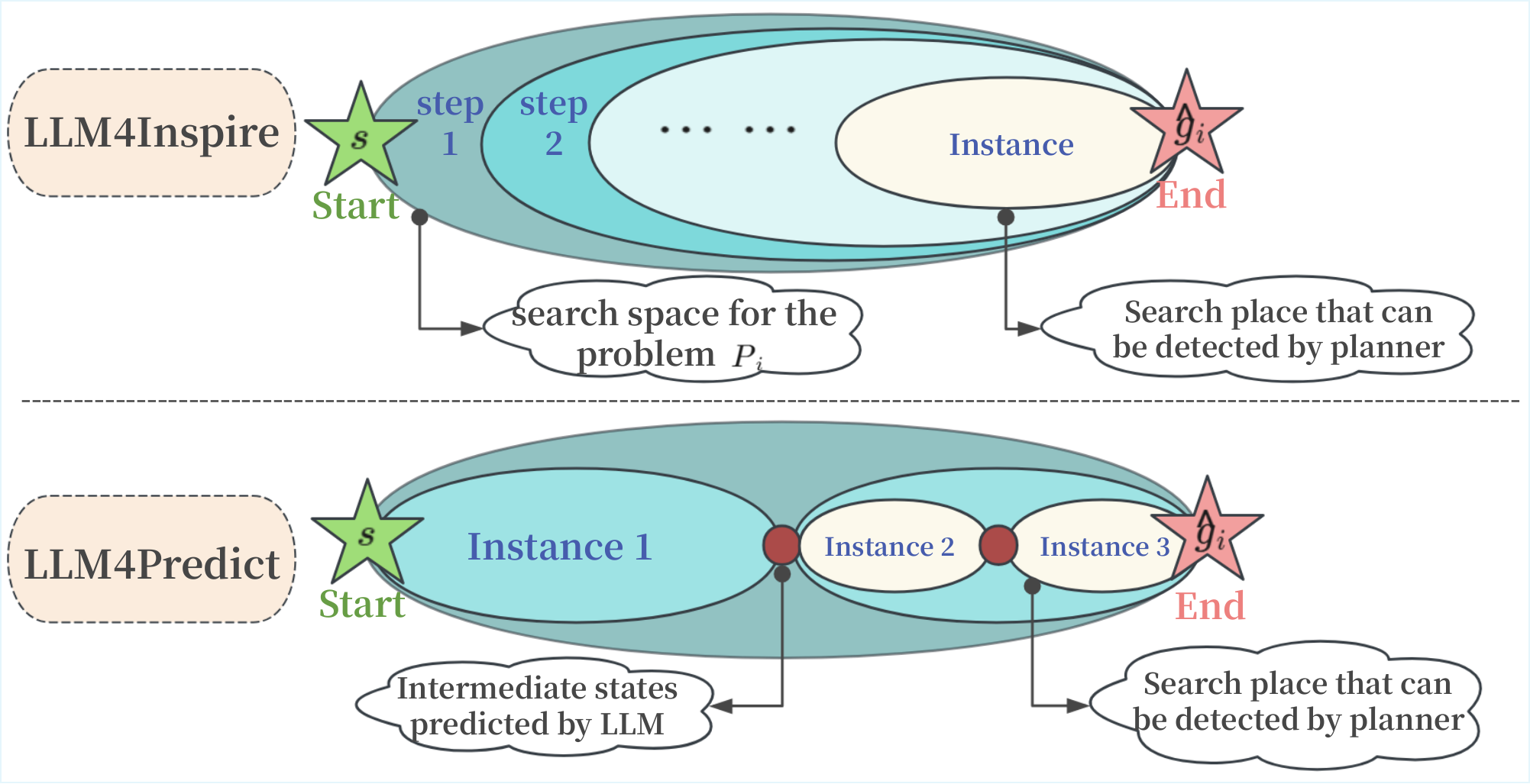}
    \caption{The overview of the impact of two LLM-assisted modes on the search space for subproblems}
    \label{fig:analysis}
\end{figure}

The candidate plan $\hat{p}$ provided by LLM4Inspire can be regarded as high-level heuristic guidance. If such a candidate plan of length $k$ is treated as a single-step guide, it is equivalent to skipping $k$ levels in the search tree, thereby reducing the problem scale, as shown in Figure \ref{fig:analysis}. Although the selected action may not be the optimal one, it is useful to prune to solve by continuously compressing the search space. 

LLM4Predict can be regarded as another novel LLM-based divide-and-conquer method. If there exists a suitable intermediate state $\tilde{s}$ that can partition the problem $P$ into two subproblems $P_1 = <s, \tilde{s}, \mathcal{D}>$ and $P_2 = <s, g, \mathcal{D}>$ where both are independently solvable, then $P$ is also solvable. The solution is the concatenation of the two sub-solutions. Formally, if $p_1$ and $p_2$ are solutions of $P_1$ and $P_2$ respectively, then $p = p_1 \circ p_2$ is necessarily a solution to $P$. This decomposition gets benefits from the shorter length of $P_1$ and $P_2$ rather than the original solution length $k$, so their respective search complexities (denoted by $O(b^{| p_1 |})$ and $O(b^{| p_2 |})$) sum to far less than the direct solution complexity $O(b^k)$. And $b$ is the average action number of each level of the search tree. Especially when $| \pi_1 | \approx | \pi_2 | \approx k/2$, the resulting complexity reduction is exponential.

\subsubsection{Soundness and Conditional Completeness}

\textbf{Theorem 1.} The action sequence computed by LLM4Inspire/LLM4Predict is a valid plan for the  planning problem $P = \langle s_0, g, D\rangle$.
% Whenever our framework returns success, the final action sequence produced by the framework is a valid executable plan for the original planning problem $P = \langle s_0, g, D\rangle$. $s_0$.

\textbf{Proof:} The framework does not directly execute LLM-generated plans. Conflict Detection first produces ordered subproblems with updated initial states and mutex constraints, and each subproblem is solved by the underlying planner. Only planner-verified sub-plans are accepted into the final solution. Thus, the final plan is a concatenation of verified executable sub-plans. Since each sub-plan is valid in its corresponding state, the returned action sequence is executable and achieves the goal. Therefore, the framework is sound. 

\textbf{Theorem 2.} If Conflict Detection successfully removes cyclic dependencies and repeated decomposition can eventually reduce the original problem $P = \langle s_0, g, D\rangle$ into finitely many subproblems that are solvable by the underlying planner, then $P$ is solvable by the framework.

\textbf{Proof:} Our framework divides a large-scale planning problem into subproblems through Conflict Detection and recursive decomposition. It does not require all subproblems from a single step to be immediately solvable. Instead, Conflict Detection removes cyclic dependencies and generates ordered subproblems with updated initial states and mutex constraints.
If a subproblem remains too complex, it can be further decomposed with LLM assistance. Assume that, given unbounded decomposition depth and planning time, this process terminates after finitely many steps and yields planner-solvable leaf subproblems. Since each leaf subproblem is solved by the underlying planner, it produces a verified executable sub-plan. Because the decomposition preserves sub-goal order and avoids interference, these sub-plans can be concatenated consistently. Therefore, the original problem can be solved by composing the verified leaf-level sub-plans. Hence, the framework is conditionally complete under recursive decomposition.

\section{Experiments}

We evaluate the performance across four domains: Blocks, Logistics, Depot, and Mystery (Round 1). The domain specifications and problem instances are from the International Planning Competitions (IPC) \footnote{https://github.com/potassco/pddl-instances}. The details of action and predicate number are shown in Table $\ref{tab:domains}$

%%%% The details of action and predicate number are ... 

1. \textbf{Blocks} requires a robot to pick up and put down blocks to achieve an initial configuration into a specified goal arrangement.

2. \textbf{Logistics} requires trucks and airplanes to transfer packages to the target locations, where trucks drive within a single city, and airplanes fly between airports.

3. \textbf{Depot} includes trucks transporting crates around. The goal is to stack crates at their destinations.

4. \textbf{Mystery (Round 1)} includes vehicles, cargo items, and some amount of fuel. The task is to load the cargo items onto vehicles and transfer them to goals with limited fuel. Note that the domain replaces all names of action, predicate, and objects with random words. 

\begin{table}[h]
    \centering
    \resizebox{0.8\textwidth}{!}{
    \begin{tabular}{c c c c c c}
    \toprule
        Domain & Action & Predicate & Objects types & Objects number & Instances  \\
        \midrule
        Blocks & 4 & 1 & 1 & 4-25 & 50\\
        Logistic & 4 & 3 & 6 & 15-51 & 42\\
        Depot & 5 & 6 & 4 & 15-72 & 22\\
        Mystery& 3 & 7 & 5 & 21-42 & 30\\
        \bottomrule
    \end{tabular}}
    \caption{Four domains used in the experiment. }
    \label{tab:domains}
    %% 补充材料
\end{table}

The Blocks and Depot domains include sequential constraints and dependencies. The Logistics domain contains a large number of objects as well as their state transitions, resulting in a large search space. The Mystery (Round 1) domain contains illogical actions, predicates, and objects, helping us critically study the ability of LLMs to guide planners. 

To investigate the planning performance, our comparative experiments include the following methods:

\textbf{Fast Downward}~\cite{helmert2006fast} is a widely used state-of-the-art classical planner in the planning community. We adopt Fast Downward with its optimal configuration (e.g., A* search with admissible heuristics) to ensure strong baseline performance. This baseline serves as a reference for evaluating the effectiveness of pure heuristic search without decomposition or LLM assistance.

\textbf{Decomposition} is a variant built upon Fast Downward, where problems are decomposed into subproblems using predefined decomposition strategies without involving LLMs. This baseline allows us to assess the impact of decomposition alone, isolating it from LLM-generated guidance.

\textbf{LLMs-only} relies solely on large language models, including GPT-5, Claude-Sonnet-4-20250514, and DeepSeek-R1, to solve planning problems. The domain description, current state, and goal are formatted into manually designed prompt templates. The LLM directly generates action sequences without invoking any classical planning algorithms. This baseline evaluates the capability of LLMs to perform planning independently.

\textbf{RAP}~\cite{hao2023reasoning} is a representative framework that integrates LLMs with planning through reasoning and planning loops. It leverages LLMs to iteratively guide the planning process and refine intermediate decisions, serving as a hybrid baseline.

\textbf{LLM4Inspire} is a variant of our framework where LLMs are used to generate executable actions directly at each decision step. This baseline evaluates the effectiveness of using LLMs as action-level guidance without explicit intermediate state prediction.

\textbf{LLM4Predict} is another variant of our framework where LLMs are employed to predict intermediate states (subgoals) to guide the planning process. This baseline reflects the contribution of LLM-based subgoal prediction without incorporating conflict-aware decomposition.

% 1. \textbf{Fast Downward  \cite{helmert2006fast}} is one of the most widely used planners in the planning community. The heuristic configuration of Fast Downward is set to the optimal path configuration in this paper.

% 2. \textbf{Decomposition} is utilized the Fast Downward to decompose the problems without LLMs. 

% 3. \textbf{LLMs-only} only uses LLMs, including GPT5, Claude-sonnet-4-20250514, and DeepSeek-R1, to solve the problems, with domains, current states, and goals formalized as hand-made prompt templates.

% 4. \textbf{RAP \cite{hao2023reasoning}} is a well-known planning framework with LLMs.

% 5. \textbf{LLM4Inspire} is configured to leverage LLMs to provide an executable action.

% 6. \textbf{LLM4Predict} is configured to leverage LLMs to predict an intermediate state.

We evaluate the approaches on the following aspects:

 \textbf{Planning success rate:} This metric evaluates valid plans generated by the planner within a \textbf{3-minute} time limit. Note that the LLMs-only method is not suitable for this cut-time limitation,  since the time taken to call LLMs can be influenced by factors, e.g., internet speed and device performance. 
% When LLMs are deployed locally, the time required to call LLMs is significantly reduced. 
Therefore, the time consumed by calling LLMs is not included.

 \textbf{Successful plan length:} We evaluate the number of steps in the valid plans. This metric reflects whether our method can balance planning success with efficiency. Typically, shorter plans indicate better performance.

 \textbf{LLMs calls and solver's time consumption:}
This metric is to explore which LLM-based method is more effective at simplifying instances.
The solver's planning time can be regarded as an indicator for evaluating the search space size, while the number of LLM calls represents the resource consumption. Fewer LLM calls indicate a stronger ability to simplify the problem. 

All experiments run with Intel(R) Core(TM) i7-14650HX (24 cores, 2.2GHz), 32,768MB of RAM, and Windows 11. Fast Downward (version 24.06.1) is deployed on Ubuntu system running on VMware. The LLM4Inspire and LLM4Predict utilize DeepSeek-R1 as LLM-based components.

\subsection{Experimental Results}
\subsubsection{Results on planning success rate}  

\renewcommand{\arraystretch}{1.2}
\begin{table}[h!]
\centering
\resizebox{0.8\textwidth}{!}{
\begin{tabular}{ccccc}
\toprule
\multirow{2}{*}{\textbf{Method}} & \multicolumn{4}{c}{\textbf{Experiment Domains}} \\ \cline{2-5}
 & \textbf{Blocks} & \textbf{Logistics} & \textbf{Depot} & \textbf{Mystery(Round-1)} \\ \midrule
Fast Downward & 26/50 & 17/42 & 5/22 & \textbf{15}/30 \\ 
Decomposition & 40/50 & 42/42 & 15/22 & \textbf{15}/30 \\ 
DeepSeek-R1 & 35/50 & 13/42 & 4/22 & 0/30 \\ 
GPT5 & 50/50 & 9/42 & 10/22 & 0/30 \\ 
Claude-sonnet-4-20250514 & 46/50 & 3/42 & 7/22 & 0/30 \\ 
RAP & 1/50 & 0/42 & 1/22 & 0/30 \\ 
LLM4Inspire & 37/50 & 42/42 & 17/22 & \textbf{15}/30 \\ 
LLM4Predict & \textbf{49}/50 & \textbf{42}/42 & \textbf{19}/22 & \textbf{15}/30 \\ \bottomrule
\end{tabular}
}
\caption{Comparison of methods across different domains.}
\label{tab:success_rate}
\end{table}

Table \ref{tab:success_rate} shows the ratio of successful cases for each method in the four fields to the total number of cases.
Compared to the other five methods, LLM4Predict performs exceptionally well across all domains, particularly achieving a success rate of over 95\% in the Blocks, Logistics, and Depot domains. 
As noted in \cite{nau2003shop2,alford2012htn}, problem decomposition is a critical reason for the outstanding performance. 
Specifically, the success rate of Fast Downward can be viewed as an indicator of domain complexity. 
The lower the success rate of Fast Downward is, the larger the corresponding search space is, indicating complex constraints between the goals. 
In the Depot domain, although LLMs-only perform poorly, the LLM-assisted method still carry out effective planning. We believe that the higher the planning success rate reflects the powerful heuristic and reasoning capabilities of LLMs. 
% Both LLM4Predict and LLM4Inspire leverage the planning capabilities along with the extensive general knowledge and domain-specific reasoning powers of LLMs. 

Note that Mystery replaces action names with random words, but it does not affect the traditional planner. However, LLMs struggle when facing domains where the logical relationships between objects and actions cannot be inferred from their literal meanings.

%-------------------------------------------------
% 在接下来这一段，1.需要解释为什么相比其他大模型我们选择采用DS——DS在各个领域预训练的嫌疑最小，并且格式校验更严格。 2.解释为什么RAP的结果那么差——MCTS作为底层规划器非常落后，这体现在这个算法需要不断地去扩展多余节点，并且每一步的跨度很小。

\textbf{Comparison between different LLMs:}
% In this paragraph, we mainly explain the reasons why DeepSeek-R1 serves as the planning component.
 During experiments, we found that although Claude demonstrates strong planning capabilities, its poor long-text processing performance leads to outputs that invariably contain redundant information when addressing complex problems. 
While GPT-5 boasts the most robust planning capabilities, it is inferior to DeepSeek-R1 for certain tasks, i.e., LLM4Inspire and LLM4Predict, in terms of output format validation.
Furthermore, since we aim to explore the planning capabilities resulting from the combination of LLMs and planners, we choose DeepSeek-R1, which has balanced results across various domains, as our LLM component.

\textbf{Analysis of the success rate of RAP:}
RAP proposes to utilize the MCTS algorithm for planning, while LLMs mainly determine the exploration value of the nodes.
This project innovatively incorporated LLM into the planning algorithm and assigned the most time-consuming task of evaluating the value of the nodes to LLM, significantly reducing the planning space of MCTS. 
However, the inherent algorithmic characteristics of MCTS determine that the planning process must consume a lot of resources to explore unknown nodes rather than those more likely ones.
This results in MCTS being unable to provide results within a limited time when dealing with complex problems, and it also consumes a large number of LLM access times. 
% Therefore, replacing MCTS with a more efficient planner, such as Fast Downward, is a more promising direction.

\subsubsection{Results on successful plan length}

\begin{figure*}[h]
    \centering
    \includegraphics[width= 1\textwidth]{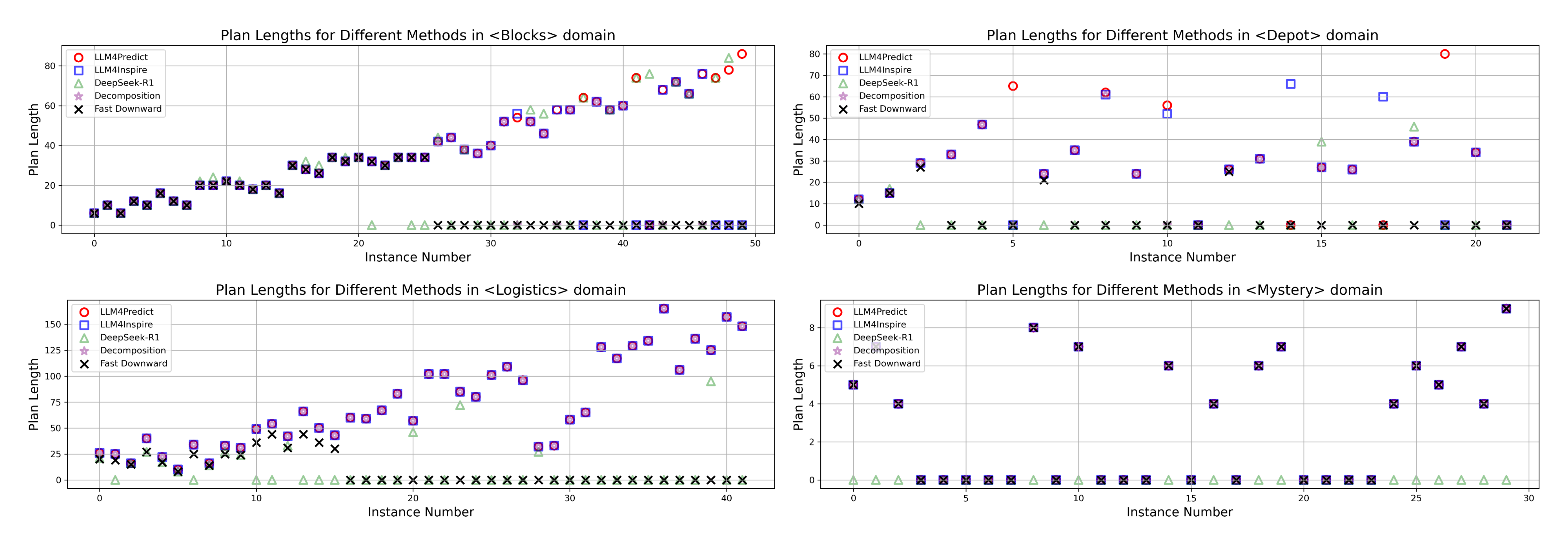}
    \caption{Plan Lengths results in the four domains}
    \label{fig:plan_len}
\end{figure*}

% %%%%%%%%%%%%%%%%%%%%%%%%%%%%%%%%%%%%%%%%%%%%%%%
% Figure \ref{fig:plan_len} presents the results on the successful plan length in four domains. In a nutshell, in the Blocks domain, DeepSeek-R1 outperformed Fast Downward on early instances, while LLM4Predict later surpassed LLM4Inspire as problem difficulty increased. In Logistics, both LLM4Predict and LLM4Inspire achieved perfect success rates, though DeepSeek-R1 required fewer planning steps when successful. 
% For Depot, LLM4Predict and LLM4Inspire solved most problems, with LLM4Predict succeeding on two instances where LLM4Inspire failed.
% In the Mystery domain, DeepSeek-R1 notably underperformed, demonstrating the overall necessity of integrating LLMs with planners rather than relying on either approach alone. The more detailed analysis has been added in Appendix A.4.
% %%%%%%%%%%%%%%%%%%%%%%%%%%%%%%%%%%%%%%%%%%%%%%%
Figure \ref{fig:plan_len} presents the results on the successful plan length in four domains. 
In the Blocks domain, where the difficulty of the instances increases as the instance number rises. All methods, except for RAP, consistently generate plans for the first 20 problems with approximately plan lengths. 
However, Fast Downward can not handle the problems after instance 26 due to an unaffordable search space.  
Next, DeepSeek-R1 performs exceptionally well in this domain. If we disregard the cut-off time, DeepSeek-R1 has already surpassed Fast Downward within the Blocks domain. Since instances between 25 and 35 can be solved using the Decomposition, LLM4Predict and LLM4Inspire perform similarly. After instance 35, the performance of LLM4Predict surpasses LLM4Inspire, indicating that in the Block domain, using LLMs to predict according to domain-specific knowledge is more advantageous than using general knowledge as heuristics.

In the Logistics domain, the overall results are slightly inferior to the Blocks domain, indirectly indicating more complex logistic relations. Impressively, both LLM4Predict and LLM4Inspire achieved 100\%, demonstrating that divide-and-conquer methods are efficient in domains with no mutual influence between predicates of goals. However, DeepSeek-R1 performs poorly. Surprisingly, the number of planning steps required for its successful plans is fewer than those of the other methods. This indirectly suggests that although the divide-and-conquer approach can solve complex problems, it does come at the cost of some planning performance.

In the Depot domain, as shown in the upper-right area of Figure \ref{fig:plan_len}, DeepSeek-R1 and Fast Downward perform poorly, 
where LLM4Predict and LLM4Inspire solve mostly problems. LLM4Predict successfully solves instances 5 and 20, where LLM4Inspire cannot, while the reverse exists for instances 15 and 18. We consider that LLM4Predict may alter previously achieved sub-goals when exploring towards the intermediate state, resulting in invalid plans when concatenating. 
Therefore, in domains where LLMs lack expertise, using LLMs to predict intermediate states for decomposition may lead to failures due to inappropriate generations provided by the LLMs.

In the Mystery domain, DeepSeek-R1 performs extremely poorly in this domain, which prevents it from providing effective guidance for both LLM4Predict and LLM4Inspire. This further demonstrates the necessity of integrating planners with LLMs, as relying solely on a planner or exclusively on LLMs is insufficient to fully address complex problems across different domains.

\subsubsection{Results on LLMs calls and Solver's time}
\begin{figure}[h]
    \centering
    \includegraphics[width=0.98\textwidth]{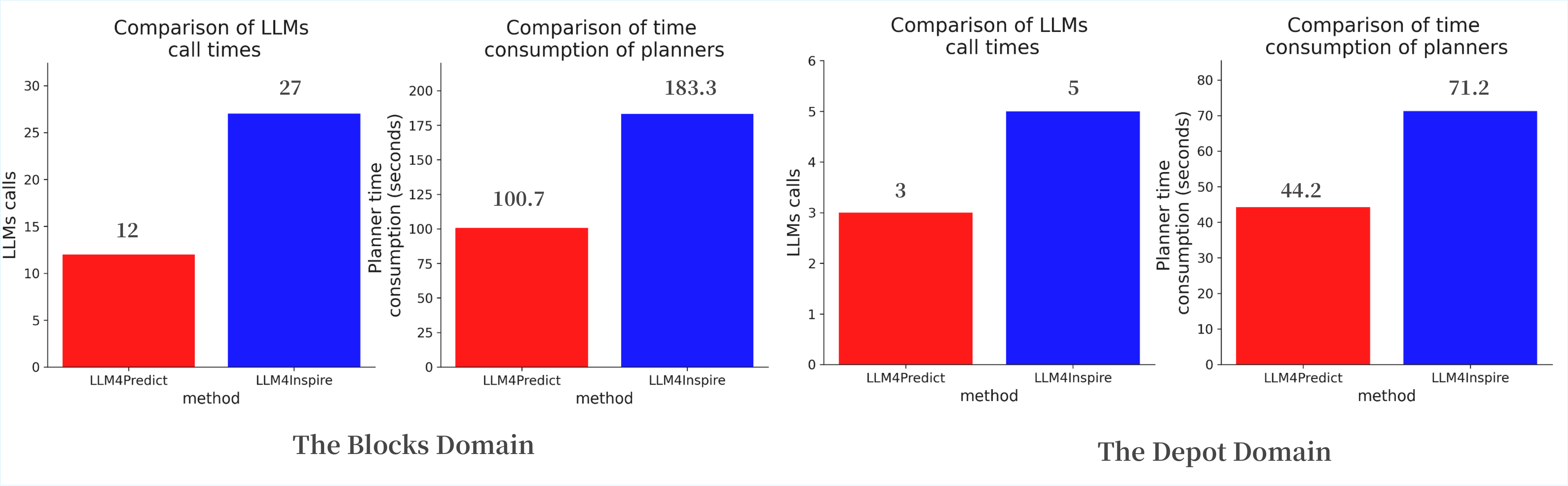}
    \caption{LLM calls and the running time.}
    \label{fig:compare}
\end{figure}

Figure \ref{fig:compare} shows the total number of LLM calls and the total planning time consumed by Solver for all instances that were successfully solved by both LLM4Predict and LLM4Inspire in the Blocks and Depot domains. Red bars represent LLM4Predict, and blue bars represent LLM4Inspire. The bar chart on the left indicates the number of LLM calls, while the right shows the planning time consumed by Solver. In both domains, LLM4Predict consistently requires fewer LLM calls and less planning time. This indicates that the LLM4Predict paradigm is more effective at pushing the problem’s search space into the planner’s solvable domain.

\subsection{Discussion on Pre-training and Data Leakage}
Although LLMs usually perform better in domains that are closer to their pre-training distribution, the effectiveness of our framework cannot be simply attributed to pre-training or data leakage.
\begin{enumerate}
    \item  The main contribution of our method lies in the conflict-aware decomposition-and-composition framework, where cyclic dependencies are detected and eliminated before the problem is decomposed into ordered subproblems for solver-based planning. Experimental results show that even the Decomposition variant, without LLM assistance, already significantly outperforms Fast Downward in multiple domains, indicating that the structural design itself contributes substantially to the final performance.

\item The capabilities of LLMs remain limited in unfamiliar domains. In the Mystery domain, where action names are replaced by random words, LLMs fail to provide reliable guidance, while the traditional planner is not affected by such renaming. This observation suggests that LLMs do not solve the problem by universally robust reasoning or hidden memorization, but rather provide domain-sensitive auxiliary guidance whose usefulness depends on their prior knowledge.

\item LLM outputs are not directly used as final plans in our framework. They are only employed to generate heuristic hints or candidate intermediate states, and every accepted result must still be verified by the symbolic planner. Therefore, the final plan quality is guaranteed by the planner, while the LLM mainly serves as a tool for reducing the search burden. 
\end{enumerate}

In a nutshell, these results indicate that the superiority of our method is not merely a by-product of LLM pre-training, but stems from the proposed integration of conflict detection, decomposition, and verified planning.

\section{Conclusion}
% In this paper, we propose a novel LLM-assisted planner integrated with problem decomposition and explore two LLM-assisted paradigms, i.e., LLM4Inspire and LLM4Predict. The experimental results have validated the effectiveness of the divide-and-conquer approach and demonstrate the capability of utilizing LLMs to handle complex tasks. In the future, we intend to explore fine-tuning LLMs to infuse them with domain-specific knowledge for large-scale planning.
In this paper, we presented a novel LLM-assisted planning framework for large-scale planning, in which LLMs are integrated into a decomposition-based symbolic planner rather than used as end-to-end planners. Based on this framework, we explored two paradigms, LLM4Inspire and LLM4Predict, to exploit LLMs for heuristic guidance during problem solving. The core idea of our method is to decompose a complex planning problem into ordered subproblems through conflict-aware analysis, and then solve them within a verifiable planning framework. Experimental results show that the divide-and-conquer strategy is effective for improving scalability, and that LLMs can further enhance performance when used as auxiliary modules. These findings suggest that combining LLMs with symbolic planning is a promising direction for solving complex problems.

In future work, we will explore fine-tuning LLMs with domain-specific data to improve their guidance ability in unfamiliar domains. We also plan to study stronger conflict-analysis and subproblem-generation strategies, as well as more adaptive ways of incorporating LLM feedback into the planning loop. In addition, extending the framework to more realistic settings, such as uncertain, dynamic, or partially observable environments, is another important direction.

% =========================
% References
% =========================
% \bibliographystyle{elsarticle-num}
\bibliographystyle{ACM-Reference-Format}
\bibliography{references}

\end{document}